\documentclass[letterpaper, 10 pt, conference]{ieeeconf} 
\IEEEoverridecommandlockouts

\pdfoutput=1
\usepackage{url}

\usepackage{graphicx}

\usepackage{amsmath,amssymb,amsfonts}

\usepackage{hyperref}

\usepackage{bm}
\usepackage{xcolor}
\usepackage{multirow}
\usepackage{paralist}
\usepackage{wrapfig}
\usepackage{cleveref}
\usepackage{siunitx}

\usepackage[style=ieee,citestyle=numeric-comp]{biblatex}
\usepackage[ruled,linesnumbered]{algorithm2e}
\usepackage[nolist,nohyperlinks]{acronym}
\begin{acronym}
\acro{MAV}{Micro Aerial Vehicle}
\newacroindefinite{MAV}{an}{a}
\acro{OMAV}{Omnidirectional Micro Aerial Vehicle}
\newacroindefinite{OMAV}{an}{an}
\acro{UAV}{Uncrewed Aerial Vehicle}
\acro{ESDF}{Euclidian Signed Distance Field}
\acro{IMU}{Inertial Measurement Unit}
\acro{RMP}{Riemannian Motion Policy}
\acro{EDF}{Euclidean Distance Field}
\newacroindefinite{IMU}{an}{an}
\acro{VTOL}{vertical take-off and landing}
\end{acronym}

\newcommand{\ray}{\vec{\mathbf{r}}}
\renewcommand{\vec}[1]{\mathbf{#1}}

\newcommand{\policy}{\mathcal{P}}
\newcommand{\robotposition}{\vec{x}}
\newcommand{\robotvelocity}{\vec{\dot{x}}}
\newcommand{\robotacceleration}{\vec{\ddot{x}}}
\newcommand{\metric}{\mathbf{A}}
\newcommand{\polf}{\vec{f}}

\newcommand{\distray}{d}

\newcommand{\nicetilde}{{\raise.17ex\hbox{$\scriptstyle\sim$}}
}
\newcommand{\Rthree}{\mathbb{R}^{3}}
\newcommand{\Rthreebthree}{\mathbb{R}^{3\times3}}
\begin{document}

\newcommand{\todo}[1]{}
\renewcommand{\todo}[1]{{\color{red} TODO: {#1}}}
\renewcommand\vec[1]{\mathbf{#1}}

\title{\emph{Waverider}: Leveraging Hierarchical, Multi-Resolution Maps\\for Efficient and Reactive Obstacle Avoidance}

\author{Victor~Reijgwart*,
        Michael~Pantic*, 
        Roland~Siegwart, 
        Lionel~Ott
\thanks{* The authors contributed equally.}%
\thanks{All authors are with the Autonomous Systems Lab, ETH Z\"urich, Switzerland %
        {\tt\footnotesize [victorr | mpantic | rsiegwart | lioott]@ethz.ch}.}%
        \thanks{This work has received funding from the EU’s Horizon 2020 programme (grant No 871542), and from Armasuisse (grant No 8003537412).}
}

\maketitle%
\thispagestyle{empty}
\pagestyle{empty}

\begin{abstract}
Fast and reliable obstacle avoidance is an important task for mobile robots. In this work, we propose an efficient reactive system that provides high-quality obstacle avoidance while running at hundreds of hertz with minimal resource usage. Our approach combines wavemap, a hierarchical volumetric map representation, with a novel hierarchical and parallelizable obstacle avoidance algorithm formulated through Riemannian Motion Policies (RMP). Leveraging multi-resolution obstacle avoidance policies, the proposed navigation system facilitates precise, low-latency (36ms), and extremely efficient obstacle avoidance with a very large perceptive radius (30m). We perform extensive statistical evaluations on indoor and outdoor maps, verifying that the proposed system compares favorably to fixed-resolution RMP variants and CHOMP. Finally, the RMP formulation allows the seamless fusion of obstacle avoidance with additional objectives, such as goal-seeking, to obtain a fully-fledged navigation system that is versatile and robust. We deploy the system on a Micro Aerial Vehicle and show how it navigates through an indoor obstacle course. Our complete implementation, called \textit{waverider}, is made available as open source\footnote{\url{https://github.com/ethz-asl/waverider}}.

\end{abstract}
\section{Introduction}
Reactive, precise, and reliable obstacle avoidance is vital for mobile robots to safely and efficiently navigate through changing or partially unknown environments.
Since obstacle avoidance is an always-on process, it must use minimal computational resources and seamlessly integrate with the robot's other tasks.
Existing approaches range from simple reactive methods using 1D distance sensors to optimization-based systems requiring complete 3D maps and vary in complexity, reaction time, and obstacle resolution.
While collision avoidance systems that operate directly on raw sensor data may exhibit exceptionally low latency, they can only guarantee safety with respect to consistently observed obstacles within the Field of View (e.g. \cite{oleynikova2015reactive}). One way to introduce memory without losing generality is to use volumetric maps. They can model obstacles of arbitrary shape and explicitly distinguish free and unobserved space. Volumetric maps are well suited to ensure safety even in unknown environments. However, fixed-resolution volumetric mapping frameworks tend to suffer from excessive memory overheads and latency. These can be overcome by using hierarchical volumetric representations such as octomap~\cite{hornung2013Octomap}, UFOMap~\cite{duberg2020ufomap}, supereight~\cite{vespa2018efficient}, or wavemap~\cite{reijgwart2023wavemap}. While several works investigated the use of hierarchical maps for global path planning, most collision avoidance systems still process all obstacles at the highest resolution. Yet, intuitively, one would expect that distant obstacles could be considered at a lower resolution than nearby ones without significantly affecting the robot's behavior.
\begin{figure}[bt]
    \centering
    \includegraphics[width=\linewidth]{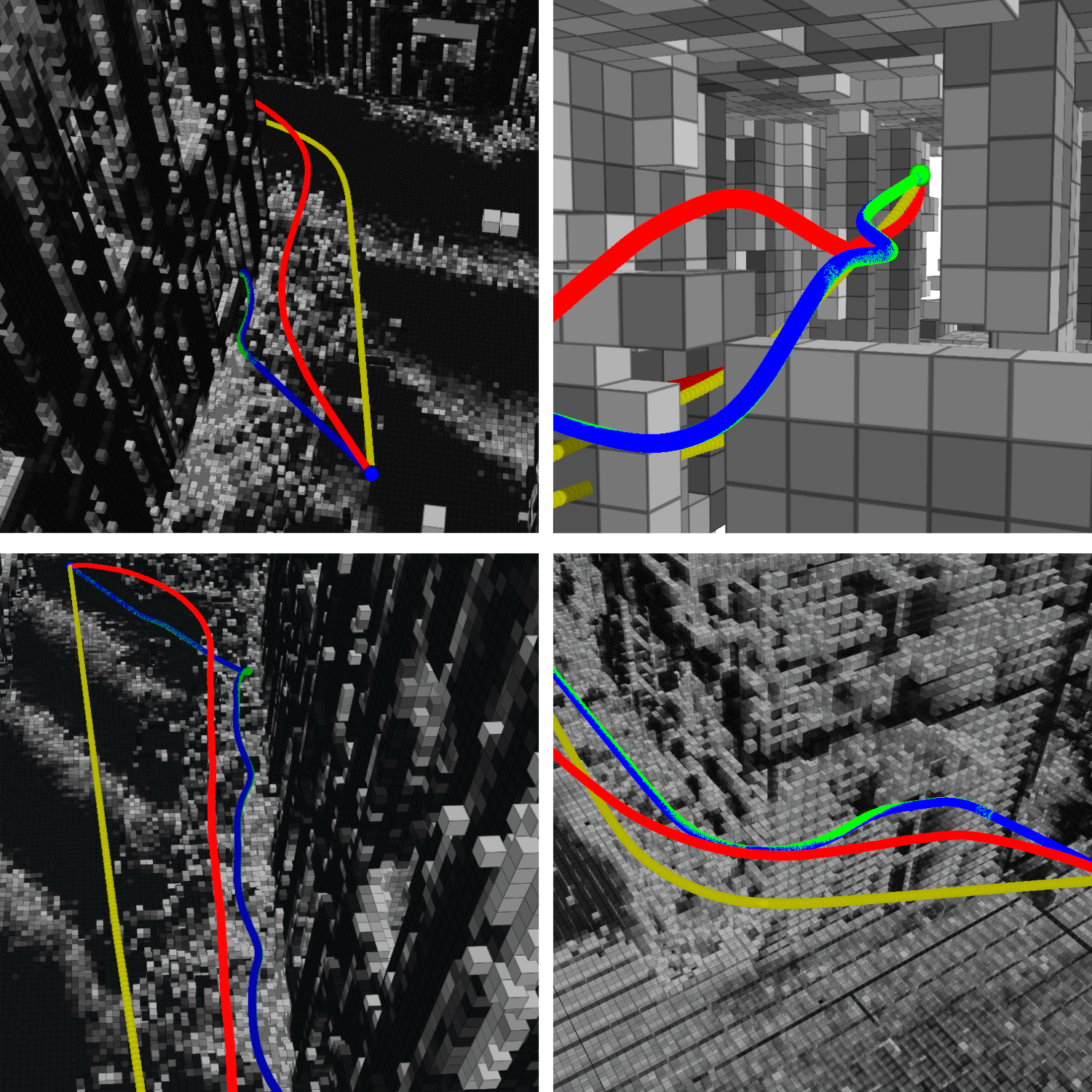}
    \caption{Example trajectories comparing our multi-resolution collision avoidance method ({\color[HTML]{dc2a0c}red}) to equivalent RMP-based formulations that consider all obstacles at the highest resolution within a radius of \SI{1}{\meter} ({\color[HTML]{90fd38}green}) and \SI{3}{\meter} ({\color[HTML]{1100fa}blue}). The fixed-resolution RMP trajectories are jerkier and more prone to get stuck (top-left). CHOMP ({\color[HTML]{bbbc28}brown}) yields smooth, albeit overly cautious trajectories and occasionally cuts through obstacles (top-right, bottom-left).}
    \label{fig:qualitative}
    \vspace{-20pt}
\end{figure}
We use \acp{RMP} \cite{ratliffRiemannianMotionPolicies2018} to formulate a navigation algorithm that is inherently multi-scale and hierarchical. \acp{RMP} are purely reactive in nature, and as such, can be formulated extremely efficiently and executed with low latency at controller frequency.
Other sampling- or optimization-based methods often need pre- and post-processing steps such as the generation of an \ac{ESDF} or trajectory smoothing. Conversely, \acp{RMP} are formulated as second-order dynamical systems and directly output accelerations, which typically leads to gradual changes and smooth paths. \acp{RMP} have some similarities to the well-known potential fields \cite{khatib1986real}, but are a much more expressive framework due to the inclusion of the Riemannian metric that modulates each policy's strength and directionality.
In this work, we develop a reactive and safe obstacle avoidance method using \acp{RMP}~\cite{ratliffRiemannianMotionPolicies2018} that is tailored to hierarchical volumetric map representations. We numerically analyze the effects of obstacle resolution on the policy's approximation error as a function of the distance between the robot and the policy. Based on this analysis, we derive a function that computes the ideal resolution for querying the map at a given distance from the robot -- allowing us to balance computational effort and accuracy.
Using this function, we develop an algorithm that efficiently generates multi-resolution avoidance policies from a hierarchical map.
The contributions of this paper are:
\begin{itemize}
    \item An efficient hierarchical obstacle policy generation algorithm;
    \item Numerical analysis of the approximation error induced by hierarchical navigation policies;
\end{itemize}
The correctness of the numerical analysis is statistically validated through a large number of experiments in simulation. Extensive comparisons with baselines and CHOMP \cite{zucker2013chomp} demonstrate the favorable run-time and efficiency of our method. Finally, we demonstrate real-world applicability by deploying our system onboard an MAV running at \SI{200}{\hertz}.
\section{Related Work}
A core decision in any obstacle avoidance system is the environment representation. State-of-the-art systems combine a volumetric map such as a truncated signed distance field \cite{oleynikova2017voxblox,han2019fiesta} or an octree-based occupancy map \cite{hornung2013Octomap,vespa2018efficient,duberg2020ufomap,reijgwart2023wavemap} with either a search-based method such as A*\cite{fastautonomousflight2018}, a sampling-based approach such as RRT \cite{karaman2011sampling}, or an optimizer such as CHOMP \cite{zucker2013chomp, oleynikova2020open}. All of these methods are comparably slow, as the mapping-planning cycle has multiple performance bottlenecks, and the sampling or optimization steps often rely on post-processed maps. Recently, end-to-end learning-based methods were shown to be effective for collision avoidance \cite{loquercio2021learning}. However, their data-driven nature still comes with a lack of generalizability across different environments, sensors, and robot dynamics.
Reactive approaches that operate directly on volumetric maps or even raw LiDAR data exist \cite{pantic2023obstacle}, but these methods have considerable memory and computing requirements due to their dense data representation. Although hierarchical volumetric maps have received considerable attention from the planning community, most works focused on global planning~\cite{Kambhampati1986OctreeAStar,Du2020MRAStar,funk2021supereight2}. Multi-resolution anytime planners~\cite{saxena2022AMRAStar} have been proposed that bridge the gap to local planning. However, their global context makes achieving the update rates required for low-latency reactive collision avoidance challenging in 3D. Nils et al.~\cite{funk2023orientationAStar} propose a full planning pipeline that leverages multi-resolution for efficient orientation-aware planning in environments with very narrow openings. However, their evaluations are performed on pre-computed static maps without perception in the loop, which makes it difficult to judge the system's latency in a reactive collision avoidance setting. Closest to our work is the hierarchical collision avoidance system presented by Goel et al.~\cite{Goel2022hierarchicalMotionPrimitives} that adapts the map resolution based on the motion primitives considered by the planner. The method is used in a teleoperation setting and shows promising results in simulated and real environments. However, a significant part of the system's efficiency results from using a bespoke, purely local map representation whose resolution is set by the planner, which is harder to reuse for additional tasks, including global planning. In comparison, our system achieves comparable efficiency levels using generic hierarchical occupancy maps. This is explained by the efficiency of RMPs, and the fact that our method does not rely on expensive \acp{ESDF}. One final benefit of our proposed architecture, compared to both~\cite{funk2023orientationAStar,Goel2022hierarchicalMotionPrimitives}, is its high degree of modularity. Formulating obstacle avoidance as a motion policy makes it easy to combine with other policies representing additional objectives such as goal-seeking, visual servoing, or aerial manipulation.
\section{Method}
\label{sec:method}
\begin{figure*}[bt]
    \centering
    \includegraphics[width=2\columnwidth]{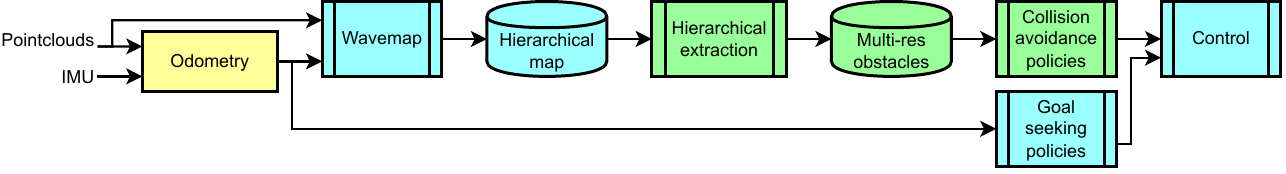}
    \caption{Block diagram of the proposed navigation system. External components are highlighted in yellow, tightly integrated components in blue, and new components introduced in this paper in green.}
    \label{fig:overview}
\end{figure*}

In the following sections, we describe our approach to efficiently extract multi-resolution obstacle avoidance policies from hierarchical maps and how they integrate with high-level task policies. 
\Cref{fig:overview} shows the main parts of the system, consisting of:
\begin{inparaenum}[1)]
    \item a volumetric, hierarchical map representation (\Cref{sec:hierarchical_map}),
    \item an algorithm for obstacle extraction (\Cref{sec:obstacle_cell_extraction}),
    \item an \ac{RMP}-based reactive navigation system (\Cref{sec:navigation_system})\end{inparaenum}.
For each obstacle cell extracted in 2) an individual obstacle avoidance policy is generated (\Cref{sec:policy_generation}), and combined with all other policies through the RMP framework. 
\subsection{Hierarchical map}
\label{sec:hierarchical_map}
The proposed method is compatible with any hierarchical occupancy mapping framework, e.g. \cite{hornung2013Octomap,vespa2018efficient,duberg2020ufomap}. We chose to use wavemap \cite{reijgwart2023wavemap}, as it simultaneously achieves state-of-the-art accuracy, memory, and computational efficiency. In a similar fashion to other methods, wavemap leverages octrees to achieve this efficiency. However, instead of storing absolute occupancy values, each node stores Haar wavelet coefficients. Using wavelets achieves significant compression and, more importantly, guarantees that all resolution levels are implicitly synchronized and always up to date. An efficient coarse-to-fine measurement integration algorithm allows wavemap to integrate depth measurements with low latency, even on computationally constrained platforms.
\subsection{Obstacle cell extraction}
\label{sec:obstacle_cell_extraction}
\begin{figure}[bt]
    \centering
    \includegraphics[width=\linewidth]{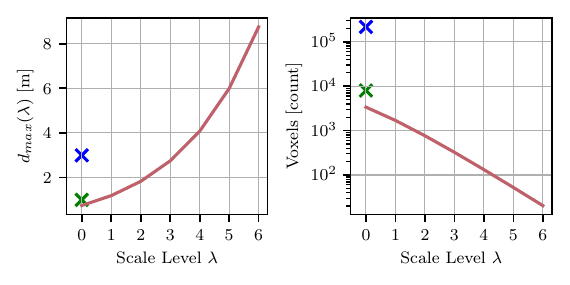}
            \vspace{-20pt}
    \caption{Left: Perceptive radius defined by $d_{max}(
    \lambda)$ as used in the obstacle filter ({\color[HTML]{bf616a}red}). Limited, fixed-resolution comparison variants used in \ref{sec:statistic_eval} are marked with a {\color[HTML]{0000ff}blue} resp. {\color[HTML]{008000}green} cross. Right: Worst-case counts of voxels to visit. Even with small perceptive radii, the fixed-resolution variants need to potentially iterate over significantly more voxels to provide the same quality of obstacle avoidance (log-scale).}
    \label{fig:perceptive_radius}
    \vspace{-20pt}
\end{figure}

As will be substantiated in \Cref{sec:proof_impact}, reducing the resolution of obstacles as the distance to the robot increases does not introduce significant approximation errors. By representing obstacles at the appropriate resolution, it is therefore possible to efficiently consider fine nearby obstacles and the broader spatial context simultaneously. In this section, we present a hierarchical algorithm that efficiently gathers multi-resolution obstacles by traversing the map in a coarse-to-fine manner. The algorithm (Algorithm \ref{algo:hierarchical_obstacle_extractor}) starts at the lowest resolution level (root node) of the map and recursively visits each node's higher-resolution children. The algorithm stops expanding a node when that node either has no children or its distance $d$ to the robot exceeds $d_{max}(\lambda)$. We use $d_{max}(\lambda)=3^{\lambda/3}-0.25$ where $\lambda$ corresponds to the node's height in the octree\footnote{A height of 0 corresponds to the highest resolution/smallest voxel size.}. Once such a terminal node is found, an obstacle cell is created if the node or any of its children is occupied.  \Cref{fig:perceptive_radius} visualizes $d_{max}(\lambda)$ and the resulting maximum distance up to which obstacles are included.
\begin{algorithm}[bt]
    \SetAlgoLined
    \SetNoFillComment
    \DontPrintSemicolon

    \scriptsize
    \KwIn{%
        Hierarchical occupancy map $\mathcal{M}$,\;
        \Indp\Indp Robot position $\vec p$
    }
    \KwOut{Set of multi-resolution obstacles $\mathcal{O}$}

    \SetKwProg{Fn}{Function}{ is}{end}

    \SetKwFunction{getOctreeRoot}{GetOctreeRoot}
    \SetKwFunction{isOccupied}{IsOcc}
    \SetKwFunction{hasOccupiedChild}{HasOccChild}
    \SetKwFunction{recursiveObstacleExtractor}{RecursiveExtractor}

    \Fn{\recursiveObstacleExtractor{$\mathcal{V}, \vec p$}}{
        $d \gets || \mathcal{V}_\text{center} - \vec p ||_2$\;
        \If{$d_{max}(\mathcal{V}_\lambda) < d$}{%
            \If{\isOccupied{$\mathcal{V}$} $\mathbf{or}$ \hasOccupiedChild{$\mathcal{V}$}}{%
                $\mathcal{O}.insert(\mathcal{V})$\;
            }
            \Return
        }

        \If{$\mathbf{not}$ \hasOccupiedChild{$\mathcal{V}$}}{%
            \Return
        }
        
        \For{$\mathcal{V}_\text{child} \in \mathcal{V}$}{%
            $\recursiveObstacleExtractor(\mathcal{V}_\text{child}, \vec p)$\;
        }
    }

    \tcp{Initialize and start recursion}
    $\mathcal{V}_\text{root} \gets \getOctreeRoot(\mathcal{M})$\;
    $\mathcal{O} \gets \recursiveObstacleExtractor(\mathcal{V}_\text{root}, \vec p)$\;

    \caption{Hierarchical obstacle extractor}
    \label{algo:hierarchical_obstacle_extractor}
\end{algorithm}

\subsection{Collision avoidance policy generation}
\label{sec:policy_generation}
\begin{figure}[bt]
    \centering
    \includegraphics[width=\linewidth]{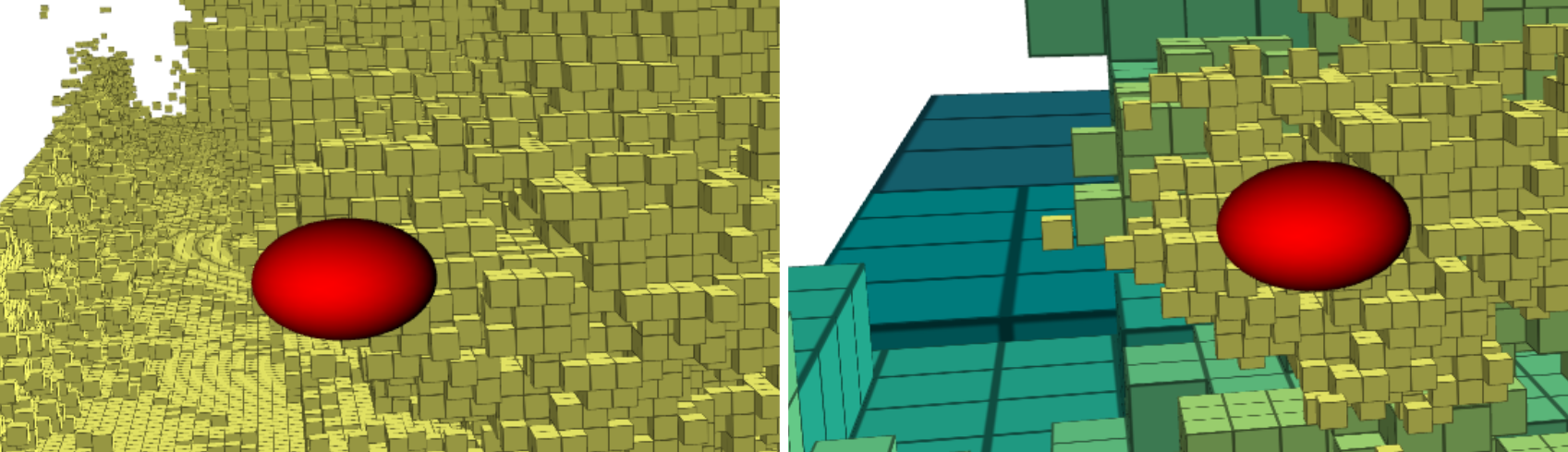}
    \caption{Comparison of an environment represented using fixed-resolution (left) and hierarchical obstacle cells (right). Our approach uses hierarchical cells, whose resolution ({\color[HTML]{969648}light brown} to {\color[HTML]{3a5d6b}dark green}) is high close to the robot ({\color[HTML]{d4280b}red}) and decreases with distance.}
    \label{fig:comparsion}
    \vspace{-15pt}
\end{figure}
For each obstacle cell returned by the previously described algorithm, an individual obstacle-avoidance policy $\policy$~\cite{ratliffRiemannianMotionPolicies2018} is created. In the following, we give a short summary of the most important aspects of motion planning using \ac{RMP}, however for more details and complete formulas of helper functions we refer to the original text ~\cite{ratliffRiemannianMotionPolicies2018}.
A policy $\policy$ consists of an acceleration function  $\robotacceleration = f(\robotposition,\robotvelocity) \in \Rthree$  and a Riemannian metric $\metric(\robotposition,\robotvelocity) \in \Rthreebthree$, where $\robotposition \in \Rthree$ refers to the robot's current position. The function $f$ drives the robot according to the policy, while the Riemannian metric $\metric$ defines a (possibly directional or isotropic) weight of the policy in comparison to other policies.
Following \cite{ratliffRiemannianMotionPolicies2018}, multiple policies $\lbrace\policy_0, \dots, \policy_N\rbrace$ can be summed into an equivalent policy $\policy_C$ using 
\begin{equation}
\mathcal{P}_c = \left(\polf_{c}, \metric_{c}\right) = \left( \left( \sum_i \metric_i \right) ^+ \sum_i \metric_i \polf_i ,\, \sum_i \metric_i \right).
\label{eq:combine_rmps}    
\end{equation}
We use the obstacle avoidance repulsor from \cite{ratliffRiemannianMotionPolicies2018} as a policy template for each found obstacle cell. It is formulated as a combinationof a pure repulsor $\polf_{rep}$, a velocity-dependent damper  $\polf_{damp}$, and a metric (weight) that becomes $0$ if the robot's velocity does not point towards the obstacle. The repulsor is defined as
\begin{equation}
    \polf_{rep}\left(\robotposition,\,\ray,\,\distray\right) = \eta_{rep}\exp \left(- \frac{d}{\upsilon_{rep}}\right) \ray \:,
\end{equation}
where $d$ is the distance to the obstacle, $\ray$ is the unit vector pointing from the obstacle to the robot, and $\eta_{rep}$ and $\upsilon_{rep}$ are tuning parameters to set the repulsor strength ($\eta_{rep}$) and scaling ($\upsilon_{rep}$).
Similarly, the damper is defined as 
\begin{equation}
    \polf_{damp}\left(\robotvelocity,\,\ray,\,\distray\right) = \eta_{damp} \bigg/ \left(\frac{\distray}{\upsilon_{damp}} + \epsilon\right) \cdot \mathbf{P}_{obs}\left(\robotvelocity, \, \ray\right) \:,
\end{equation}
again with $\eta_{damp}$ as a strength parameter and $\upsilon_{damp}$ as a scaling parameter. $\epsilon$ is a sufficiently small constant to ensure numerical stability. $\mathbf{P}_{obs}\left(\robotvelocity, \, \ray\right)$ projects the robot velocity onto the direction vector pointing from the obstacle to the robot and captures how much the robot moves towards the obstacle.
Finally, the full obstacle avoidance policy is defined as the tuple $\policy_{obs} = (\polf_{obs}, \metric_{obs})$:
\begin{align}
    \label{eq:pol}
    \polf_{obs}\left(\robotposition, \, \robotvelocity, \, \ray,\, \distray\right) &= \polf_{rep}\left(\robotposition, \, \ray,\, \distray \right) - \polf_{damp}\left(\robotvelocity, \, \ray,\, \distray \right) \\
    \label{eq:metric}
    \metric_{obs}\left(\robotposition, \, \robotvelocity, \, \ray,\, \distray\right) &=  w_r\left(\distray, \right) \cdot \bm{s}\left(\polf_{obs}\right)
    \bm{s}\left(\polf_{obs}\right) ^T.  
\end{align}
$\bm{s}\left(\cdot\right)$ is a soft-normalization function. Please refer to \cite{ratliffRiemannianMotionPolicies2018} for the detailed formulations of $\mathbf{P}_{obs}$ (eq. 68) and $\bm{s}$ (eq. 24). $w_r$  scales the policy response based on a distance parameter $r$, which influences the policy's maximum active range according to $w_r\left(d\right) = \frac{1}{r^2}d^2 - \frac{2}{r} d + 1$.
For each of the thousands of found obstacle cells such a policy is created. All cells at the same scale level $\lambda$ are then summed according to \cref{eq:combine_rmps}, and all resulting combined policies of all scales are then again summed using \cref{eq:combine_rmps}. The scale level $\lambda$ is used to set the RMP's parameters as follows: $\upsilon_{damp} = 0.45\lambda$, $\upsilon_{rep} = 0.75\lambda$, and $r=1.5\lambda$. Modulating $\upsilon_{damp}$, $\upsilon_{rep}$, and $r$, allows setting the sphere of influence of policies, and for example determines the traversability of narrow corridors. By using the tuning proposed above, coarse obstacles naturally have a larger sphere of influence.
The distance and size of the obstacle cell are used to scale the policy's Riemannian metric, which can be interpreted as a multi-dimensional weight and modulates the policy's strength and activation radius. The Riemannian metric ensures that the relative direction to the obstacle cell is taken into account such that there is only a repulsion component if the robot's velocity points towards this obstacle. In \Cref{fig:comparsion}, examples of obstacle cells are shown for both uniform and hierarchical cell generation.

\subsection{Navigation system integration}
\label{sec:navigation_system}

We use the simple goal-attractor policy described in \cite{ratliffRiemannianMotionPolicies2018} to combine the previously described summation of obstacle avoidance policies with goal-seeking behavior. The goal-attractor policy is defined as:
\begin{equation}
\label{eq:attractorPolicy}
\begin{aligned}
    \polf_a(\robotposition, \robotvelocity) &= \alpha_a \bm{s} \left( \robotposition_a - \robotposition \right) - \beta_a \robotvelocity \\
    \metric_a(\robotposition, \robotvelocity) &= \mathbb{I}^{3 \times 3}
    \end{aligned}\;,
\end{equation}
where $\alpha_a, \beta_a > 0$ are tuning parameters, and $\robotposition_a$ is the desired goal location.
In each iteration, all policies are evaluated, summed up, and the resulting acceleration executed on the robot. For simulation experiments, the policies are run as fast as possible, whereas during field tests the policies are evaluated at the robot's control frequency (\SI{200}{\hertz}).
Note that it is straightforward to replace or combine the goal-seeking policy with other policies addressing tasks such as visual servoing, terrain following, manipulation, or assisted manual control, as has been shown e.g. in \cite{mattamala2022reactive}.
\section{Hierarchical Policy Approximation Error}
\label{sec:proof_impact}
\begin{figure}[bt]
    \centering
    \includegraphics[width=.9\linewidth]{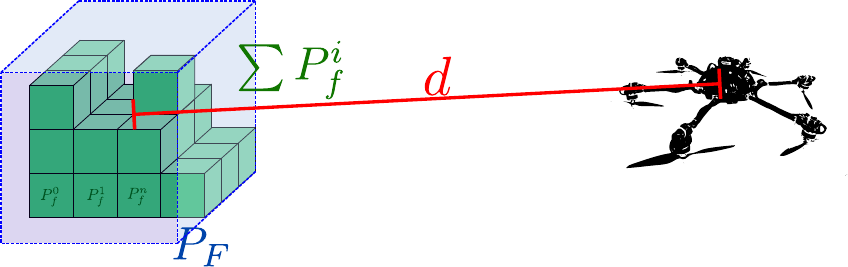}
    \caption{Example of obstacles that can be either modeled by a single, large policy ($P_{F}$) or multiple small, high-resolution policies ($P^i_f$). The distance $d$ represents the distance from the robot to the center of the obstacle block.}
    \label{fig:compression_effects}
    \vspace{-10pt}
\end{figure}

\begin{figure}[bt]
    \centering
    \includegraphics[width=\linewidth]{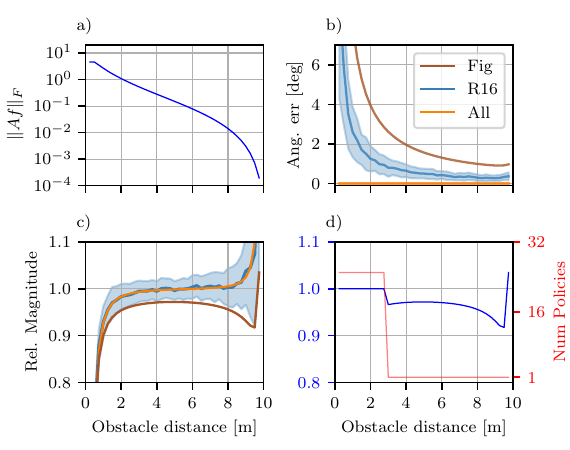}
    \vspace{-10pt}
    \caption{a) Plot of the typical absolute policy strength (log-plot) w.r.t. obstacle distance. Subfigure b) and c) visualize the angular and relative magnitude error of approximating the high-resolution policies with a single coarse approximation. `Fig' shows this for the exact configuration seen in \cref{fig:compression_effects},  `R16' for a random selection of 16 occupied voxels at high resolution (thus the covariance), and `All' for a fully occupied volume. d) Illustration of the approximation error for a hierarchical policy, where a full-resolution policy is used below \SI{2.5}{\meter} distance and a single summary policy at larger distances. The spike in the approximation error's magnitude at a distance of \SI{10}{\meter} is unimportant, as the absolute strength at this distance nears $0$.}
    \label{fig:compression_effects_eval}
    \vspace{-15pt}
\end{figure}

Naturally, one wonders what the impact of incorporating distant obstacles at a reduced resolution is. In this section, we study the influence of replacing a sum of obstacle avoidance policies with a single policy at the center of such a block. In the obstacle cell extraction algorithm, the octree is traversed to a deeper or shallower level depending on the distance to the robot. This implies that at larger distances, fewer policies at slightly different locations contribute to the overall navigation result instead of a sum of many individual policies. In the following, we show what relative changes in policy outputs and quality these abstractions entail, using the toy example in \Cref{fig:compression_effects} for the analysis.
We conduct a numerical analysis to simulate the relative changes between the single simplified policy $P_{F}$ and the granular, high-resolution set of policies $\sum P^i_f$ in both policy strength and directionality for three scenarios: 
\begin{inparaenum}[1)]
    \item the toy example in \Cref{fig:compression_effects} (labeled ``Fig'' in \Cref{fig:compression_effects_eval}),
    \item a random sampling of 16 occupied voxels, respectively their resulting policies (``R16"), and
    \item a completely occupied block resulting in 64 policies (``All'').
\end{inparaenum}
The same $4 \times 4 \times 4$ block with $\SI{10}{\centi\meter}$ voxels is used in all scenarios.
As is visible in \Cref{fig:compression_effects_eval}, the induced errors are negligible both in angular error as well as relative strength (magnitude) of the resulting policies. As to be expected, errors are higher when the distance to the voxels is smaller. The combination of multiple policies at different scales provides the best compromise; it minimizes the number of policies needed while also providing low approximation error over the entire distance.
\section{Experiments}
We perform a comprehensive set of experiments to evaluate the navigation success rates, computational efficiency, and real-world applicability of the proposed system. To provide context, we include comparisons with CHOMP~\cite{zucker2013chomp}. CHOMP generates complete trajectories and requires an \ac{EDF}, which is time-consuming to generate ($\approx$ \SI{30}{\second} for the maps used). By contrast, an RMP-based navigation framework is inherently reactive and only needs obstacle information which is readily available in the volumetric map.
\subsection{Statistical evaluation and comparison}
\label{sec:statistic_eval}
Despite the purely reactive nature of the proposed system, we are interested in its capability and performance in finding moderately complex trajectories in realistic maps. To this end, we perform an in-depth randomized evaluation on maps generated from the \textit{Newer College} LiDAR dataset \cite{zhangMultiCameraLiDARInertial2022} with a min voxel size of \SI{10}{\centi\meter}.
\begin{figure}[bt]
    \centering
    \includegraphics[width=\linewidth]{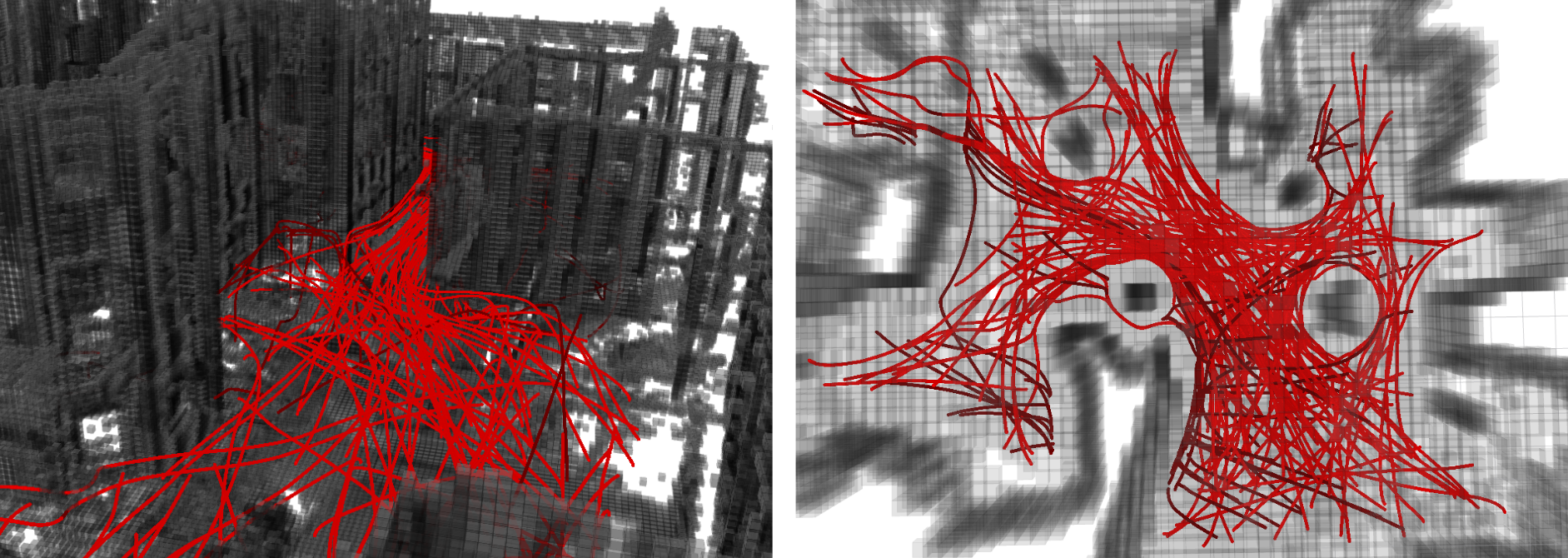}
    \vspace{-15pt}
    \caption{Qualitative visualization of the map scenarios used for statistical evaluation. Left shows a perspective rendering of the \texttt{math} scenario, right is a top-down rendering of the \texttt{mine} scenario. The red lines are example trajectories from our proposed navigation algorithm. Trajectories stuck in local minima are marked in dark red.}
    \label{fig:statistic_experiment}
    \vspace{-10pt}
\end{figure}
We use subsections of two maps -- \texttt{mine} and \texttt{math}, visualized in \Cref{fig:statistic_experiment}~-- in which we sample random start and end points and let each navigation algorithm find a smooth, collision-free trajectory. We evaluate a total of four algorithms; \begin{inparaenum}[a)]
    \item the proposed hierarchical system as described in \cref{sec:method},
    \item an implementation of CHOMP \cite{zucker2013chomp},
    \item a non-hierarchical variant of our system that only uses the highest resolution voxels, up to a maximum distance of \SI{1}{\meter}, and
    \item \SI{3}{\meter}, respectively\end{inparaenum}.
The non-hierarchical variants serve to illustrate the effects of the reduced perceptive radius, which is limited due to significantly increased compute costs inherent to single-resolution approaches at small voxel sizes. All reactive, \ac{RMP}-based variants are used in an end-to-end fashion, meaning that the policies are repeatedly updated and integrated until the robot is at a stand-still, either at the goal or in a local minima. Obstacle cells are updated from the map whenever the displacement since the last update exceeds $\SI{0.05}{\meter}$. CHOMP is configured to run with $N=500$ trajectory points until it converges ($\epsilon_{rel} < 1e^{-5}$) or a maximum iteration count (\num{100}) is reached.
\begin{figure}[bt]
    \centering
\includegraphics[width=\linewidth]{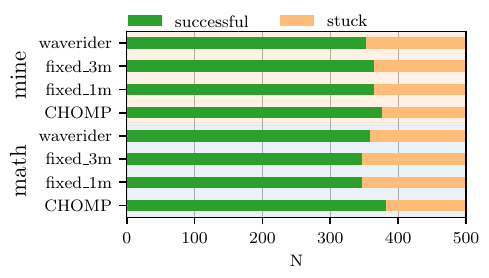}
    \vspace{-20pt}
    \caption{Success rates for all algorithms on both maps with 500 randomized trials each. CHOMP runs that did not terminate within the allocated time budget are labeled as stuck.}
    \label{fig:benchmark_success}
        \vspace{-10pt}

\end{figure}
To demonstrate the relative performance of the proposed system, we provide a detailed look at planning success rate, planning time, and distances to obstacles.
\Cref{fig:benchmark_success} shows the relative amount of successfully found trajectories, i.e. that reach the goal location and do not get stuck. All algorithms perform similarly well and solve about $75\%$ of all tasks, which is rather good considering that they are all local and not global planners. Due to their different nature, the reactive algorithms get (safely) stuck in local minima, whereas the optimization-based CHOMP method may simply not converge to a solution that is collision free.
\begin{figure}[bt]
    \centering
    \includegraphics[width=\linewidth]{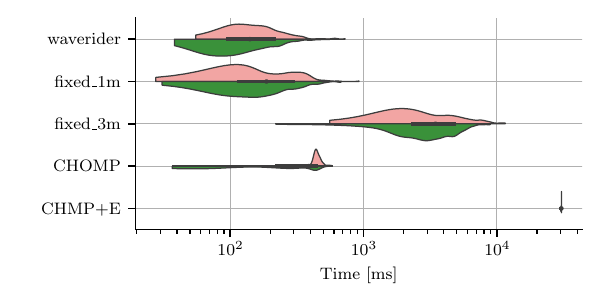}
    \vspace{-20pt}
    \caption{Timing distributions for rest-to-rest trajectories on map \texttt{math}. Green parts only include successful trajectories, red parts only stuck ones. CHOMP clearly shows increased calculation time for failing trajectories as it runs more solver iterations. Note the log scale and the drastically increased run-time for the fixed-resolution variant. For context, \texttt{CHMP+E} visualizes the cost of a trajectory, including the necessary collision distance (\ac{EDF}) pre-processing for a map for CHOMP.}
    \label{fig:timings}
    \vspace{-10pt}
\end{figure}
\Cref{fig:timings} provides detailed statistics of the measured run-times of the different algorithms. Noteworthy is the drastic increase in run-time with larger perceptive radii, which makes the use of large amounts of occupancy information intractable when a fixed-resolution representation is used. A major difference between our proposed method and CHOMP is its pure reactive nature. While we compare full rest-to-rest trajectory run-times in \Cref{fig:timings}, in practice only a single iteration is calculated at every controller iteration. Effectively, this provides full obstacle avoidance navigation at a marginal compute cost -- approximately \SI{100}{\micro\second} per step on average -- and a few milliseconds per step involving obstacle updates. Conversely, CHOMP only provides results after full convergence.
\begin{figure}[bt]
    \centering
    \includegraphics[width=\linewidth]{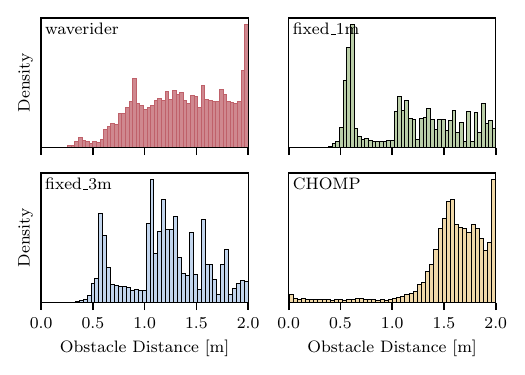}
    \vspace{-20pt}
    \caption{Histogramms of distance to obstacles over \num{50000} random trajectory traces from each algorithm. The \ac{EDF} used in the evaluation is truncated at \SI{2}{\meter}, with everything above that value being considered far away from obstacles.}
    \label{fig:histogramm}
    \vspace{-15pt}
\end{figure}
To provide insights into trajectory safety, we evaluate the distance to the closest occupied obstacle for each step along each evaluated trajectory from the randomized tests. The resulting distributions are visualized as histograms in \Cref{fig:histogramm}. The proposed hierarchical approach shows a safe distribution with no parts of the trajectories getting close to obstacles. The two fixed-resolution algorithms frequently travel \textit{much} closer to obstacles due to their limited perceptive fields, whereas CHOMP may output unsafe states in case of non-convergence.
Finally, \Cref{fig:qualitative} shows a visualization of trajectories generated in four example scenarios. These examples show that both fixed-resolution variants produce poor and unsteady trajectories due to their limited perceptive range. Combining all the presented results, the proposed multi-resolution, purely reactive, hierarchical algorithm provides an attractive compromise of run-time, success rate, and trajectory quality at marginal compute cost.
\subsection{Computational efficiency}
\begin{figure}[bt]
    \centering
    \includegraphics[width=0.85\linewidth]{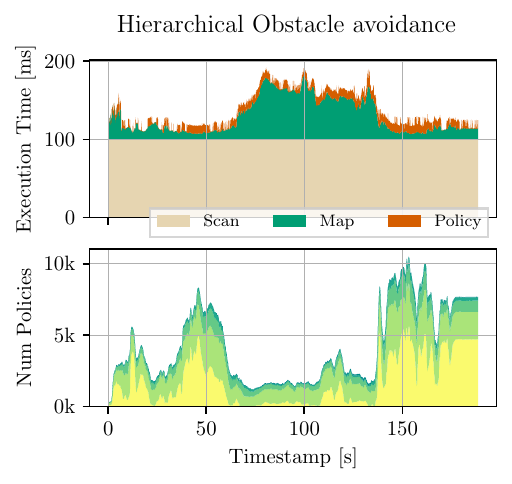}
    \vspace{-10pt}
    \caption{Top: Stackplot of data latency (LiDAR) and processing latency (Map/Policy). Bottom: Visualization of the number of policies at different levels, where yellow is the finest resolution and dark green is the coarsest, in similar fashion to \Cref{fig:comparsion}. The system is on the outdoor terrace between $\SI{55}{\second}-\SI{130}{\second}$. Especially after the robot reenters the building, it is in close proximity to many obstacles, leading to more policies at a higher resolution.}
    \label{fig:benchmark}
\end{figure}

We benchmark the computational efficiency of the proposed navigation system on a NVidia Jetson Orin AGX computer, using data from a Livox Mid-360 LiDAR.
The navigation algorithm only uses the computer's 12-core ARM Cortex-A78AE CPU. \Cref{fig:benchmark} visualizes the latency and policy processing times on a dataset that traverses indoor offices before transitioning to a terrace, including a large $\SI{30}{\meter}$ radius open space. Together, the mapping and planning use $2.4$ CPU threads (average) and $\SI{355}{\mega\byte}$ of RAM (peak). The LiDAR delivers new data every $\SI{100}{\milli\second}$ and integrating these observations takes $\SI{29}{\milli\second}$ (average), while selecting and executing the obstacle avoidance policies takes $\SI{6.9}{\milli\second}$ (average). All together, the mapping and planning steps are completed almost instantaneously after the LiDAR data is received.
\subsection{Platform tests}
\begin{figure}[bt]
    \centering
    \includegraphics[width=\linewidth]{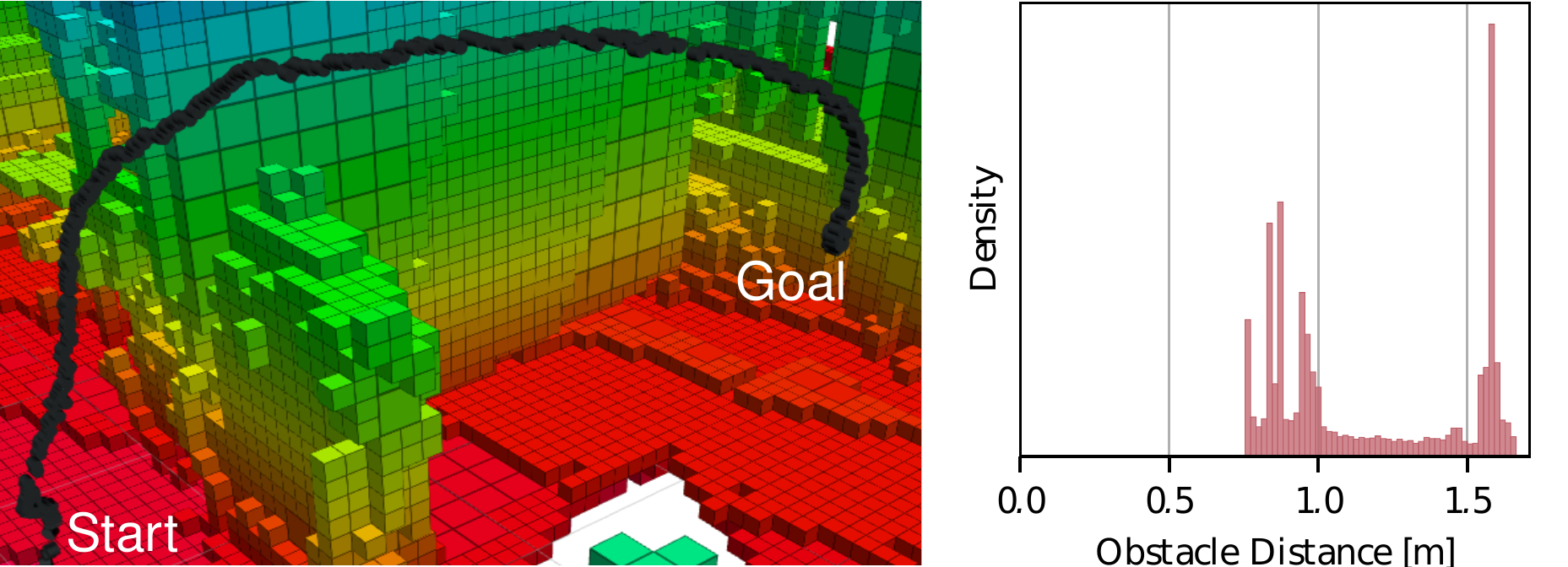}
    \caption{Left: Rendering of an executed flight path (black) and the map that was created during a traversal of a cluttered region with the specified goal location. Right: Obstacle distance histogram for the same flight. The MAV successfully cleared all obstacles with sufficient margin. Note: For operational reasons, the tuning for the field test was more conservative (stronger) than for the map-based evaluation.}
    \label{fig:fieldtest}
    \vspace{-15pt}
\end{figure}

The proposed navigation pipeline (\Cref{fig:overview}) is deployed on an \ac{MAV} with a Livox Mid-360 LiDAR for odometry \cite{xu2022fast} and mapping. We run the aerial robot through an indoor obstacle course without a prior map, such that all data used for navigation must be gathered and processed on the fly. The operator sets a desired goal location prior to the flight, which the robot then autonomously tries to reach using the proposed reactive navigation algorithm.
\Cref{fig:fieldtest} visualizes a typical path taken by the aerial robot to avoid an obstacle and fly towards a (potentially unreachable) goal position. Upon setting a desired goal position, the goal-seeking policy starts to drive the robot. After about $\SI{130}{\milli\second}$, the first scan is received, the map is populated and the obstacle avoidance policies become active. As can be seen from \Cref{fig:fieldtest}, the robot avoids the obstacles with sufficient distance. During the full run, the robot never got closer than $\SI{0.75}{\meter}$ to an obstacle and kept an average closest-obstacle distance of $\SI{1.16}{\meter} \pm \SI{0.32}{\meter}$.  
\section{Conclusion}
In this paper, we presented a novel method for multi-resolution, reactive obstacle avoidance in generic 3D environments. A key insight is the efficient use of multi-resolution, hierarchical obstacle information. This follows the intuition that geometry further away does not need to be incorporated at the same resolution as nearby obstacles. As demonstrated through numerical analysis and ablations, the proposed approach enables locally precise and safe collision avoidance while keeping a very large perceptive radius. Multi-resolution obstacles can efficiently be extracted by directly exploiting the hierarchical structure present in hierarchical volumetric mapping frameworks such as wavemap \cite{reijgwart2023wavemap}. The proposed system achieves planning success rates comparable to CHOMP while reducing the planning time by $50\times$ and requiring no pre-processing or post-processing steps, such as \ac{EDF} generation and trajectory tracking control. Finally, the system is deployed on a real MAV negotiating an indoor obstacle course while only using minimal computational resources.
\printbibliography
\end{document}